\documentclass[conference, a4paper]{IEEEtran}
\usepackage{cite}
\usepackage{amsmath,amssymb,amsfonts}
\usepackage{algorithmic}
\usepackage{graphicx}
\usepackage{textcomp}
\usepackage[T1]{fontenc}
\usepackage[utf8]{inputenc} 
\usepackage{paralist}
\usepackage{graphics} 
\usepackage{epsfig} 
\usepackage{epstopdf}
\usepackage[colorlinks=true,bookmarks=false]{hyperref}
\usepackage{subfig} 
\usepackage{lipsum}  
\AtBeginDocument{%
  \renewcommand\tablename{TABLO}
}

\AtBeginDocument{%
  \renewcommand\abstractname{Abstract}
}

\begin{document}

\title{Kısa Konuşma Cümlelerinin Doğal Dil İşleme Yöntemlerini Kullanarak  Otomatik Etiketlenmesi \\
Auto-tagging of Short Conversational Sentences using Natural Language Processing Methods}

\author{\IEEEauthorblockN{Şükrü Ozan\IEEEauthorrefmark{1} and D. Emre Taşar\IEEEauthorrefmark{1}}
\IEEEauthorblockA{\IEEEauthorrefmark{1}AdresGezgini A.Ş. Ar-Ge Merkezi, {\footnotesize İ}zmir, Türkiye\\  
    sukruozan@adresgezgini.com, emretasar@adresgezgini.com}
}
\IEEEoverridecommandlockouts
\maketitle
\IEEEpubidadjcol

\begin{ozet}

Bu çalışmada, belirli bir alana özgü cümlelerin otomatik olarak etiketlenmesi için bir yöntem bulmayı hedeflemekteyiz. Eğitim veri setimiz, bir firmanın müşteri temsilcileriyle, internet sitesi ziyaretçileri arasında gerçekleşen sohbet görüşmelerinden elde edilen kısa konuşma cümlelerinden oluşmaktadır. Yaklaşık 14 bin adet ziyaretçi girişini teker teker etiketleyerek, daha sonra anlamlı bir diyalog üretebilecek bir sohbet robotu (chatbot)  uygulaması geliştirmek amacıyla kullanılacak olan dönüştürücü tabanlı, dikkat mekanizmalı bir dil modelinde kullanılmak üzere on temel kategoriye ayırdık. Üç farklı güncel modeli ele alarak  otomatik etiketleme kabiliyetlerini karşılaştırdık. En iyi sonucu transformatörlerden çift yönlü kodlayıcı gösterimleri mimarileri (BERT)  ile elde ettik. Deneylerde kullanılan model uygulamaları GitHub depomuzdan klonlanarak benzer otomatik etiketleme problemleri için kolaylıkla test edilebilir.
\end{ozet}
\begin{IEEEanahtar}
otomatik etiketleme, doğal dil işleme, doc2vec, LSTM, BERT.
\end{IEEEanahtar}

\begin{abstract}
In this study, we aim to find a method to auto-tag  sentences specific to a domain. Our training data comprises short conversational sentences extracted from chat conversations between company's customer representatives and web site visitors. We manually tagged approximately 14 thousand visitor inputs into ten basic categories, which will later be used in a transformer-based language model with attention mechanisms for the ultimate goal of developing a chatbot application that can produce meaningful dialogue. We considered three different state-of-the-art models and reported their auto-tagging capabilities. We achieved the best performance with the bidirectional encoder representation from transformers (BERT)  model. Implementation of the models used in these experiments can be cloned from our GitHub repository and tested for similar auto-tagging problems without much effort. 
\end{abstract}
\IEEEpeerreviewmaketitle
\IEEEpubidadjcol

\section{G{\footnotesize İ}r{\footnotesize İ}ş}\label{sec:giris}

Son yıllarda, derin öğrenme literatüründe olduğu gibi, onun bir alt başlığı olan doğal dil işleme (NLP) alanında da meydana gelmekte olan gelişmeler sayesinde, dilden dile çeviri, anlamsal analiz  ve metin sınıflandırma gibi problemlerin çözümünde önemli ilerlemeler kaydedilmiştir.

Mikolov vd. tarafından geliştirilen word2vec yöntemi \cite{Miokolov:2013} ile kelimeler cümle içerisinde diğer kelimelerle birlikte kullanılma sıraları ve olasılıklarını temsil eden ve kelime gömme olarak adlandırılan çok boyutlu sayısal yöneyler şeklinde gösterilebilmektedir. \cite{Kilimci:2019} ve \cite{Sel:2019} gibi çalışmalarda kelime gömme tabanlı metin sınıflandırma yöntemleri önerilmiştir.

Yine Mikolov vd. tarafından, word2vec'e paralel olarak geliştirilen doc2vec yöntemi \cite{Mikolov:2014} ile cümle ve paragrafları da bir yöney şeklinde gösterebilmek mümkün olabilmektedir. Doc2vec, bir sınıflandırma yöntemi ile birlikte, bir doküman sınıflandırma modeli olarak da kullanılabilmektedir. 

Tang vd., \cite{Tang:2015} çalışmasında Twitter’dan alınmış cümlelerin sınıflandırılmasında uzun-kısa vadeli bellek (LSTM) modelinin kullanılmasının sözdizimsel ayrıştırıcılardan ve harici duyarlılık sözlüklerinden daha iyi performans gösterdiğini ortaya koymuşlardır.

Ayata vd., \cite{Ayata:2017} çalışmasında 2017 Twitter’da yazılmış metinlerin, kelime gömüleri modeli ile oluşturulan vektörlerin, destek vektör makinesi ve rastgele orman sınıflandırma modelleri kullanılarak sektörel bazda sınıflandırılması çalışmasını gerçekleştirmişlerdir.

Ertuğrul vd., \cite{Ertugrul:2018} çalışmasında metin verilerinden oluşan film konu özetleri ile film sınıflandırması yapmak için, cümlelere ait gömme yöneyleriyle ,çift yönlü LSTM (bi-LSTM) modelini eğiterek elde ettikleri sonuçların, tekrarlayan sinir ağları ve lojistik regresyon ile karşılaştırıldığında bu modelin daha iyi olduğunu göstermişlerdir.

Çiftçi vd., \cite{Ciftci:2018} çalışmasında Lojistik Regresyon ve Naive Bayes gibi analiz yöntemleri ile LSTM yöntemini, Türkçe duygu analizi alanında karşılaştırmış ve LSTM yönteminin daha iyi sonuç verdiğini göstermişlerdir.

Geçtiğimiz bir kaç sene içerisinde, çok daha güncel bir model olan BERT \cite{Devlin:2018} doğal dil işlemede farklı problemlerin çözümünde çok tercih edilen bir dil modeli olmuştur. BERT modelinin duygu analizi için kullanımına yönelik \cite{Gao:2019} ve \cite{Acikalin:2020} gibi güncel çalışmalar mevcuttur. 

Biz de bu çalışmada AdresGezgini A.Ş. firmasının (kısaca firma olarak adlandırabiliriz) internet sitesini ziyaret eden ziyaretçilerle firmanın müşteri temsilcileri arasında gerçekleşen  sohbet yazışmalarını ele aldık. Bu yazışmalarda sayfa ziyaretçilerinin gerçekleştirdiği sorguları anlamlandırabilmek adına bu cümleleri  manuel olarak etiketleyerek yukarıda örneklerine değindiğimiz doc2vec, LSTM ve BERT ile kurguladığımız üç adet sınıflandırma modelini bu veri ile eğittik. Doc2Vec ve LSTM modellerini, çok daha güncel bir yöntem olan  BERT modeli ile kıyasladık.

\section{Kullanılan Ver{\footnotesize İ} Set{\footnotesize İ}} \label{sec:data}

2014 ve 2020 yılları arasında firma web sitesindeki sohbet arayüzü yardımı ile müşteri temsilcileri ve site ziyaretçileri arasında gerçekleştirilen sohbet yazışmaları detaylı bir şekilde incelenerek düzgün diyalog formatında ilerleyen 13794 adet sohbet görüşmesi belirlendi. Daha sonra, bu görüşmelerde ziyaretçiler tarafından gerçekleştirilen girdi cümleler ayrıştırılarak ön tanımlı 10 adet kategori ile manuel olarak etiketlendi. Dijital reklam alanında hizmet veren firmamıza gelen sorular genellikle farklı platformlarda reklam vermek, web sitesi tasarımı yaptırmak, seo v.b. dijital reklamcılık hizmetleri ile ilgili olmaktadır. Kategorilerle etiketlediğimiz bazı örnek cümleler Tablo \ref{tablo:ornekler}'de görülebilir.

\begin{table}[h!]
  \centering
  \caption{\textsc{Örnek cümleler ve eşleşt{\footnotesize İ}r{\footnotesize İ}lm{\footnotesize İ}ş et{\footnotesize İ}ketler}}
  \label{tablo:ornekler}
  \begin{tabular}{|l|l|}
    \hline
    \textbf{Etiket} & \textbf{Örnek Cümle}\\
    \hline
    \hline
    Kategori 0  & sitemin trafiğini arttırmak için reklam vermeyi düşünüyorum\\              & internet üzerinden reklamlar vermek istiyorum\\
                & farklı yerlerden nasıl reklam verebiliyoruz\\
    \hline
    Kategori 1  & bizim internet sayfamız yok nasıl web adresi verebilirim\\                 
                & şirketim için web sitesi yaptırmak istiyorum\\ 
                & web tasarımı yaptıracağım yardımcı olabilir misiniz\\
    \hline
    Kategori 2  & google şirket reklamı vermek istiyordum \\
                & google üzerinden reklam vereceğim\\ 
                & arama sonuçlarında üst sırada çıkmak istiyorum\\
    \hline
    Kategori 3  & facebook reklamlarını kullanmak istiyoruz\\
                & şirketim için facebookta tanıtım yapacağım\\ 
                & facebook reklam vermek istiyorum\\
    \hline
    Kategori 4  & mağazamın ınstagram hesabı için reklam vermeyi düşünüyorum\\               & instagram reklamları hakkında bilgi alacağım\\
                & instagramda nasıl tanıtım yapabiliriz\\
    \hline
    Kategori 5  & youtube üst sırada çıkmak için reklam vermek istiyorum\\
                & youtube reklam nasıl oluyor\\
                & internetten youtube a reklam açacağım\\
    \hline
    Kategori 6  & seo hizmetiniz hakkında bilgi verebilir misiniz\\ 
                & siteme seo yaptıracağım\\ 
                & parayla seo yapıyor musunuz \\
    \hline
    Kategori 7  & benimle iletişime geçer misiniz\\ 
                & müşteri temsilcisiyle görüşeceğim\\
                & beni arar mısınz hemen\\
    \hline
    Kategori 8  & iyi günler dükkanımı haritaya kaydetmek istiyorum\\ 
                & google harita kaydı nasıl yapılıyor\\
                & işletme harita kaydı oluşturacağım\\
    \hline
    Kategori 9  & ücretlendirme politikanız hakkında bilgi almak istiyordum\\                & fiyatlar hakkında bilgi alabilir miyim\\
                & reklam ücretleriniz ne kadar\\
    \hline
  \end{tabular}
\end{table}

\begin{figure}
	\centering
		\includegraphics[width=0.50\textwidth]{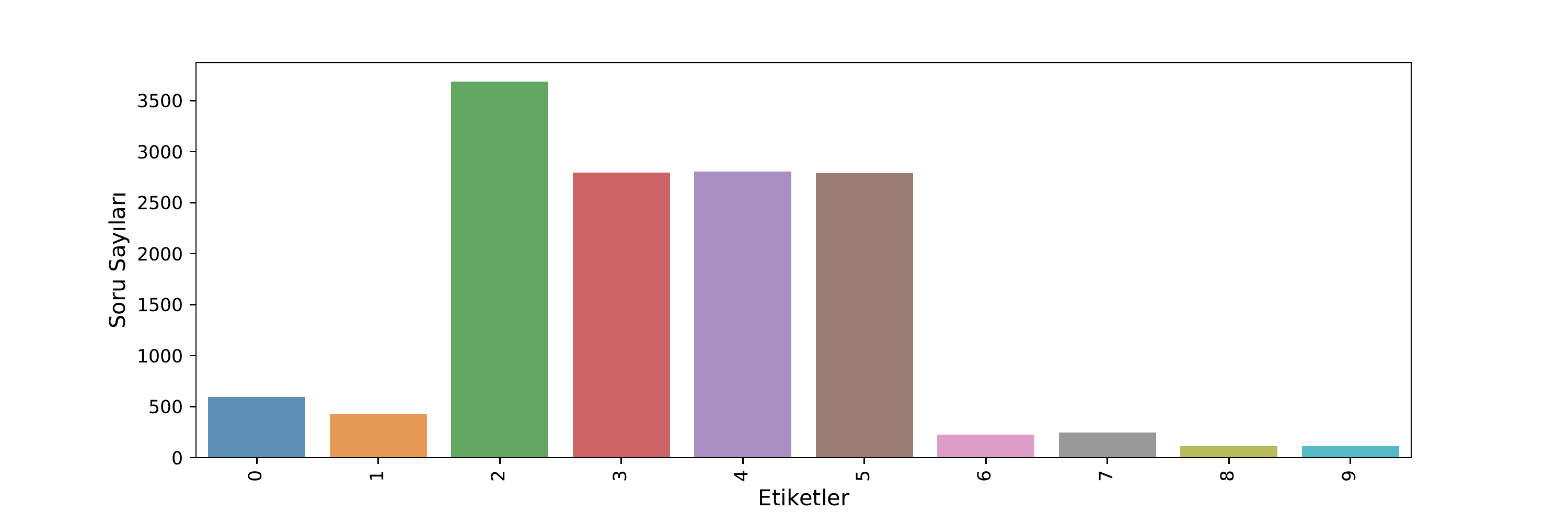}
	\caption{Veri setindeki örneklerin etiketlere göre dağılım grafiği. Grafikten de görüleceği üzere veri dağılımı büyük oranda düzensizdir.}
	\label{img:dataDist}
\end{figure}

Etiketlere göre elimizdeki verinin dağılımı Şekil \ref{img:dataDist}'de görülmektedir. Şekilden de görüleceği üzere, veri setinin etiketlere göre dağılımı büyük oranda düzensiz bir yapıdadır. Toplam veri setinin \%30 luk kısmını test için ayırarak, 9651 adet eğitim verisi veri ile bir sonraki bölümde detaylarını vereceğimiz yöntemleri inceledik.

\section{Öner{\footnotesize İ}len Yöntemler} \label{sec:methods}

\subsection{Doc2Vec Yöntemi} \label{subsec:doc2vec}

Mikolov'un\cite{Mikolov:2014} çalışmasında önerilen doc2vec yöntemi, verilen bir yazı külliyatı kullanılarak, yazı külliyatı içerisindeki kelimelerin, cümlelerin, paragrafların bir araya gelme şekilleri gözetilerek gerçekleştirilen olasılık hesapları, verilen külliyat içerisindeki yazı örneklerini, tıpkı kelime gömülerinde \cite{Miokolov:2013} olduğu gibi çok boyutlu sayısal bir yöney şeklinde gösterebilmemizi sağlamaktadır.

Çalışmamızda biz eğitim veri kümesini kullanarak bir doc2vec model  eğitimi gerçekleştirdik.  Sonuç olarak her cümle örneği için elde ettiğimiz kelime gömmelerini de   çok kategorili  yapısal bağıntı (multi nominal logistic regression (MNLR)) sınıflandırma modelini eğitmek için kullandık.  

Doc2vec modeli eğitilmeden önce veri seti noktalama işaretlerinden arındırılmış, büyük harfler küçük harfe dönüştürülmüş, NLTK kütüphanesi kullanılarak etkisiz kelimeler (stop words) kelimeler atılmıştır. İlgili veri seti ile birlikte kullanıldığında bu modelin başarımının düşük olduğu gözlemlenmiştir.

\subsection{LSTM Yöntemi} \label{subsec:lstm}

İlk olarak Hochreiter vd. tarafından \cite{Hochreiter:1997} çalışmasında önerilen uzun-kısa vadeli bellek (LSTM) mimarisi, verilerdeki zamansal bağıntıların öğrenilmesi açısından önemli bir işleve sahiptir. 

Bu model kullanılırken, veri setimizde yaygın kullanılan 6872 farklı kelime belirlenmiş ve her biri bir tamsayı ile eşleştirilmiştir (tokenization). Veri setindeki  en uzun cümlenin kelime sayısını (250) göz önünde bulundurularak daha kısa cümleleri $<BOŞ>$ ($<PAD>$) etiketi le işaretleyerek, 250 uzunluğunda eğitim girdileri elde ettik. Eğitim girdilerimizdeki sayıların ifade ettiği kelimeleri 100 boyutunda bir gömme yöneyine dönüştürecek olan gömme katmanını bir seyreltme (dropout) katmanı ardından LSTM katmanına bağlayarak bu mimariyi, girdilerimize karşılık gelen etiket bilgisi ile birlikte eğittik. Yaklaşık 5 milyon parametreden oluşan bu mimarinin başarımının bir sonraki bölümde açıklayacağımız model olan BERT modelinden daha kötü fakat doc2vec modelinden daha iyi olduğu gözlemlenmiştir.

\subsection{BERT Yöntemi} \label{subsec:bert}
Devlin vd.\cite{Devlin:2018} çalışması ile doğal dil işleme literatürüne yakın bir zamanda giren dönüştürücü (transformer) tabanlı bir dil modeli olan BERT, farklı problemlerin çözümü için önceki modeller yerine tercih edilmeye başlamıştır.  Biz de çalışmamız kapsamında gerçekleştirmeye çalıştığımız sınıflandırma problemi için Bölüm \ref{subsec:doc2vec} ve \ref{subsec:lstm}'de sırasıyla anlattığımız Doc2Vec ve LSTM yöntemleri ile birlikte BERT modelinin de başarımını test ettik. BERT ile kurulan çözüm yöntemleri önceden eğitilmiş bir model gerektirdiği için bu yaklaşım öğrenme aktarması (transfer-learning) olarak düşünülebilir. Bu duruma istinaden biz de, Türkçe dilinde önceden eğitilmiş olan  BERT-BASE-TURKISH-UNCASED modelini \cite{Schweter:2020}  tercih ettik. Bu model yaklaşık 35GB'lık oldukça büyük bir Türkçe külliyat ile eğitilmiştir. Modelin Türkçe dilinde NLP alanında başarılı sonuçlar alan ince ayar yapılmış (fine-tuned) modelleri bulunmaktadır. Biz de kendi veri setimiz ile gerçekleştirdiğimiz bir ince ayar aşamasından sonra modeli kendi sınıflandırma problemimiz için denedik. Bir sonraki bölümde de göstereceğimiz üzere en iyi test başarımını da bu model ile elde ettik.

\section{Uygulama Sonuçları}

\begin{figure*}[h!]
	\centering
		\subfloat[][Doc2Vec Yönteminin Hata Dizeyi]{\includegraphics[width=0.32\textwidth]{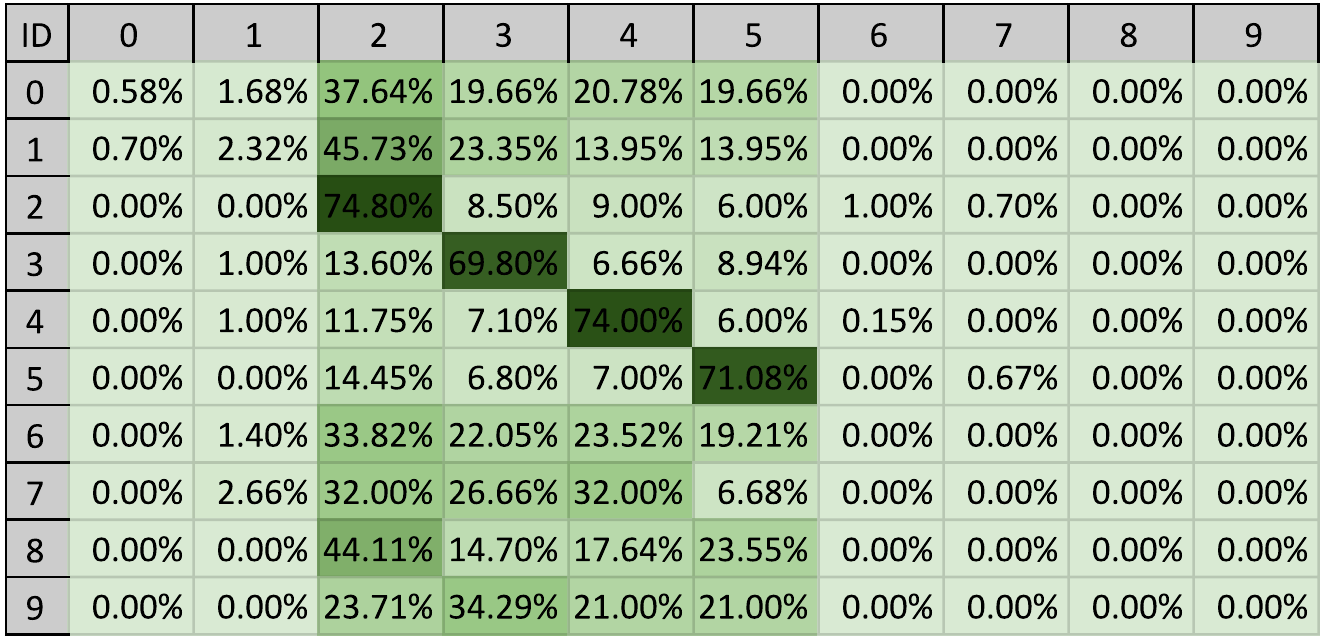}\label{img:confmxdoc2vec}}
       	\subfloat[][LSTM Yönteminin Hata Dizeyi]{\includegraphics[width=0.32\textwidth]{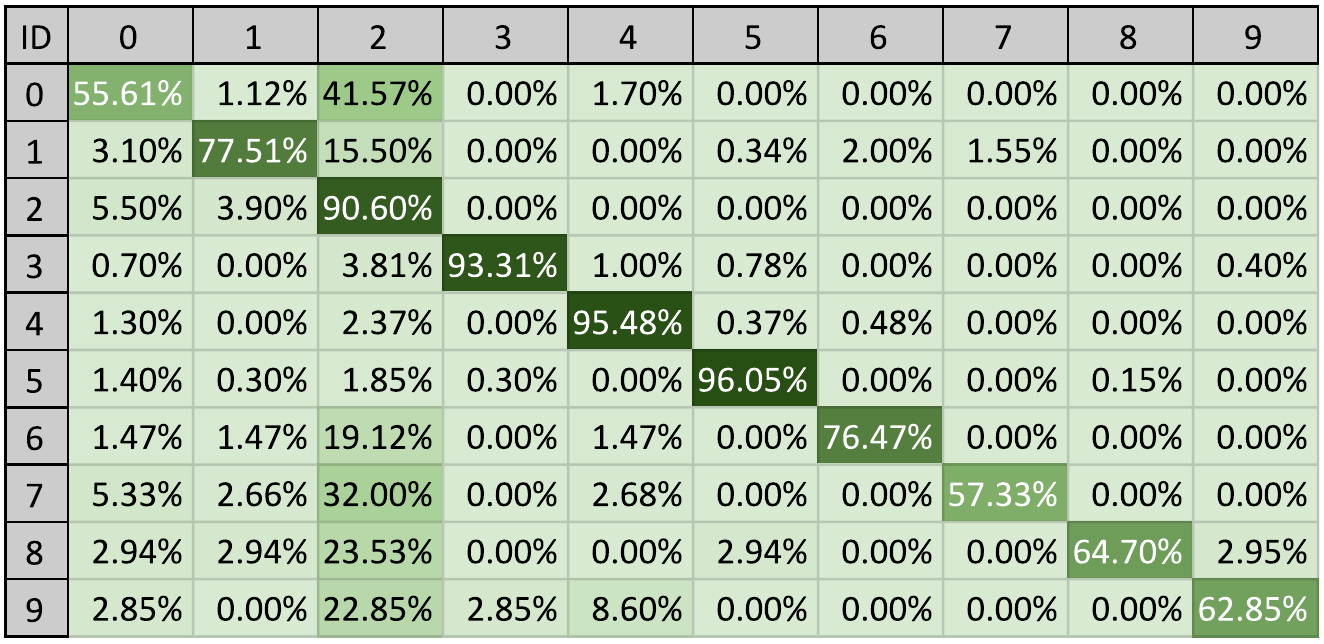}\label{img:confmxlstm}}
       	\subfloat[][BERT Yönteminin Hata Dizeyi]{\includegraphics[width=0.32\textwidth]{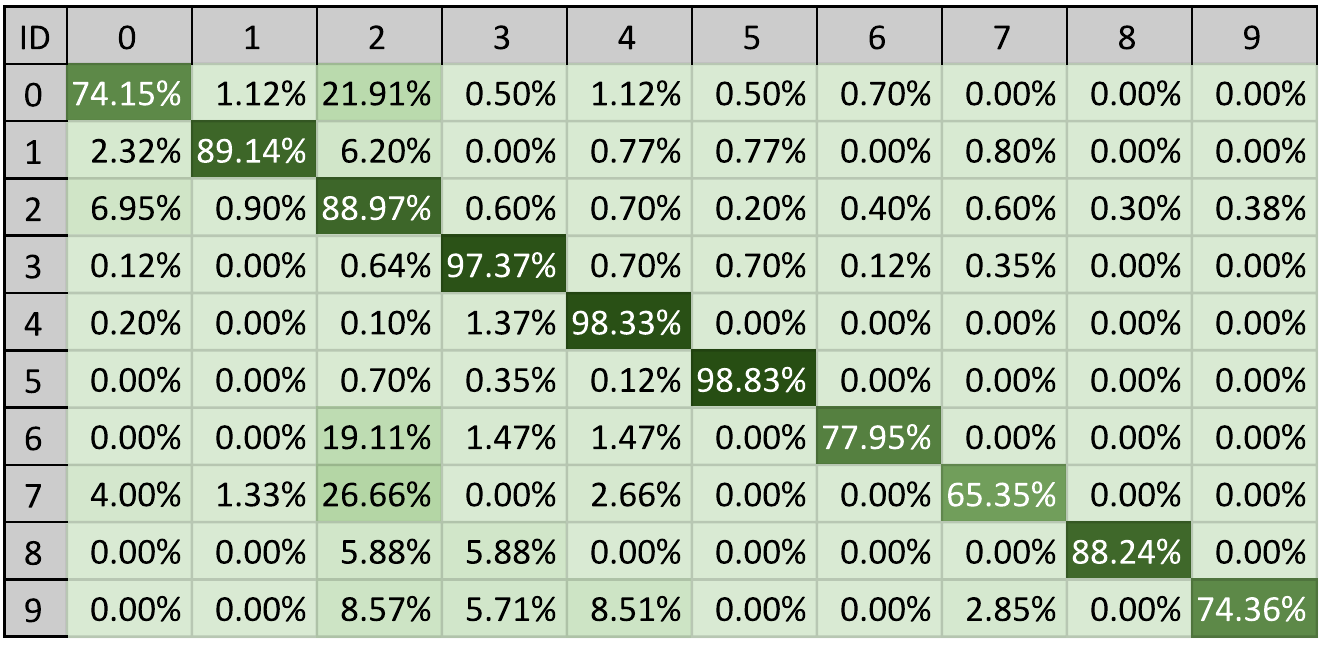}\label{img:confmxbert}}        	
       	     	
	\caption{Bölüm \ref{sec:methods}'te anlattığımız modellere ait hata dizeylerinin renklendirilmiş gösterimleri. Koyu renkler daha iyi başarım yüzdelerini temsil etmektedir. \protect\subref{img:confmxdoc2vec} Bölüm \ref{subsec:doc2vec}'da anlatılan Doc2Vec yönteminin hata dizeyi. \protect\subref{img:confmxlstm} Bölüm \ref{subsec:lstm}'de anlatılan LSTM yönteminin hata dizeyi.
	\protect\subref{img:confmxbert} Bölüm \ref{subsec:bert}'de anlatılan BERT yönteminin hata dizeyi.
	} 
	\label{img:confmxs}
\end{figure*}

Bölüm \ref{sec:giris}'de tanımını yaptığımız sınıflandırma problemini, Bölüm \ref{sec:methods}'de anlattığımız üç yöntem ile çözmeye çalıştık. Bu bölümde denediğimiz yöntemlerin sınıflandırma başarımlarını karşılaştıracağız. Yöntemlerin sınıflandırma başarımlarını Şekil \ref{img:confmxs}'de hata dizeyi (matrisi) ve Tablo \ref{tablo:yontemler}'de  tablo olarak gösterdik. Başarım hesaplarını, eğitim ve test veri seti için ayrı ayrı gerçekleştirdik. Bununla birlikte kategorik sınıflandırma problemlerinde başarımı ölçmek için kullanılan F1 metriğini de, her etiket sınıfı için ayrı ayrı hesapladık. Her ne kadar, ikili sınıflandırma yöntemlerinin başarımlarını göstermek için tek bir F1 skoru hesaplanması yeterli olsa da, buradaki gibi çoklu sınıflandırma durumlarında, F1 skorunun her bir sınıf için ayrı ayrı hesaplanması gerekmektedir. Bu durumda genel bir F1 skoru hesaplamak için iki temel yöntem kullanılmaktadır. Şayet, veri setindeki örnek sayısı her bir grup için yaklaşık eşit düzeylerde ise her bir sınıf için hesaplanmış olan F1 skorlarının aritmetik ortalaması bir metrik olarak kullanılabilmektedir, Tablo \ref{tablo:yontemler}'de bu skoru ortalama F1 skoru olarak gösterdik. Ancak buradaki gibi örnek sayısının dağılımının düzensiz olduğı veri setleri için her etiket sınıfı için hesaplanan F1 skorunun örnek sayısı baz alınarak hesaplanan bir ağırlıklı ortalamasını almak çok daha doğru bir sonuç vermektedir. Tablomuzda bu skoru ağırlıklı ortalama F1 skoru olarak gösterdik.

\begin{table*}[h!]
\centering
  \caption{\textsc{Öner{\footnotesize İ}len Yöntemler{\footnotesize İ}n Doğruluk Karşılaştırması}}
  \label{tablo:yontemler}
  \begin{tabular}{|l|c|c|c|c|}
    \hline
    \textbf{Yöntem} & \textbf{Eğitim Doğruluğu} & \textbf{Test Doğruluğu} & \textbf{Ort.F1 Skoru} & \textbf{Ağ.Ort.F1 Skoru}\\
    \hline
    \hline
    Doc2Vec + MNLR (Bölüm \ref{subsec:doc2vec}) & 0.67 & 0.64 & 0.60 & 0.28\\
    \hline
    LSTM (Bölüm \ref{subsec:lstm}) & 0.97 & 0.90 & 0.90 & 0.80\\
    \hline
    BERT (Bölüm \ref{subsec:bert}) & \textbf{0.97} & \textbf{0.93} & \textbf{0.93} & \textbf{0.87} \\
    \hline
  \end{tabular}
\end{table*}

Doc2Vec yöntemi, her ne kadar uzun cümlelerin ve paragrafların anlamsal olarak sınıflandırılmasında başarılı bir yöntem olsa da, buradaki gibi  kısa konuşma cümlelerinin sınıflandırılmasında düşük performans göstermiştir. Yöntemin eğitim ve test veri setleri için hesaplanan doğruluk oranları yüzde 60 dolaylarında çıkmıştır. Yine benzer şekilde F1 skorlarının, özellikle ağırlıklı F1 skoru değerinin çok düşük olduğu gözlemlenmiştir.

\begin{figure*}
	\centering
		\subfloat[][LSTM Yönteminin Kayıp Eğrisi]{\includegraphics[width=0.45\textwidth]{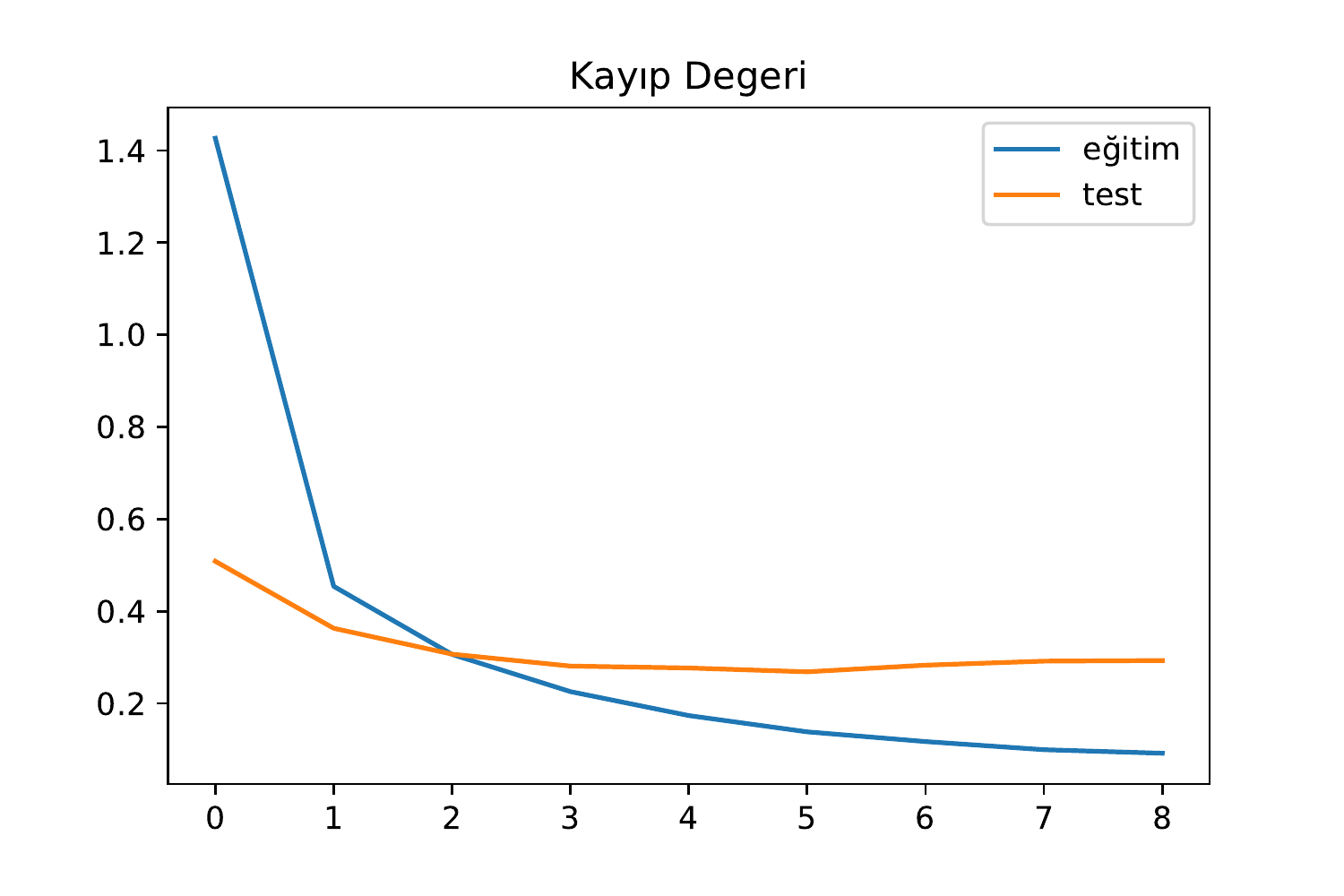}\label{img:lstmLoss}}
       	\subfloat[][LSTM Yönteminin Doğruluk Eğrisi]{\includegraphics[width=0.45\textwidth]{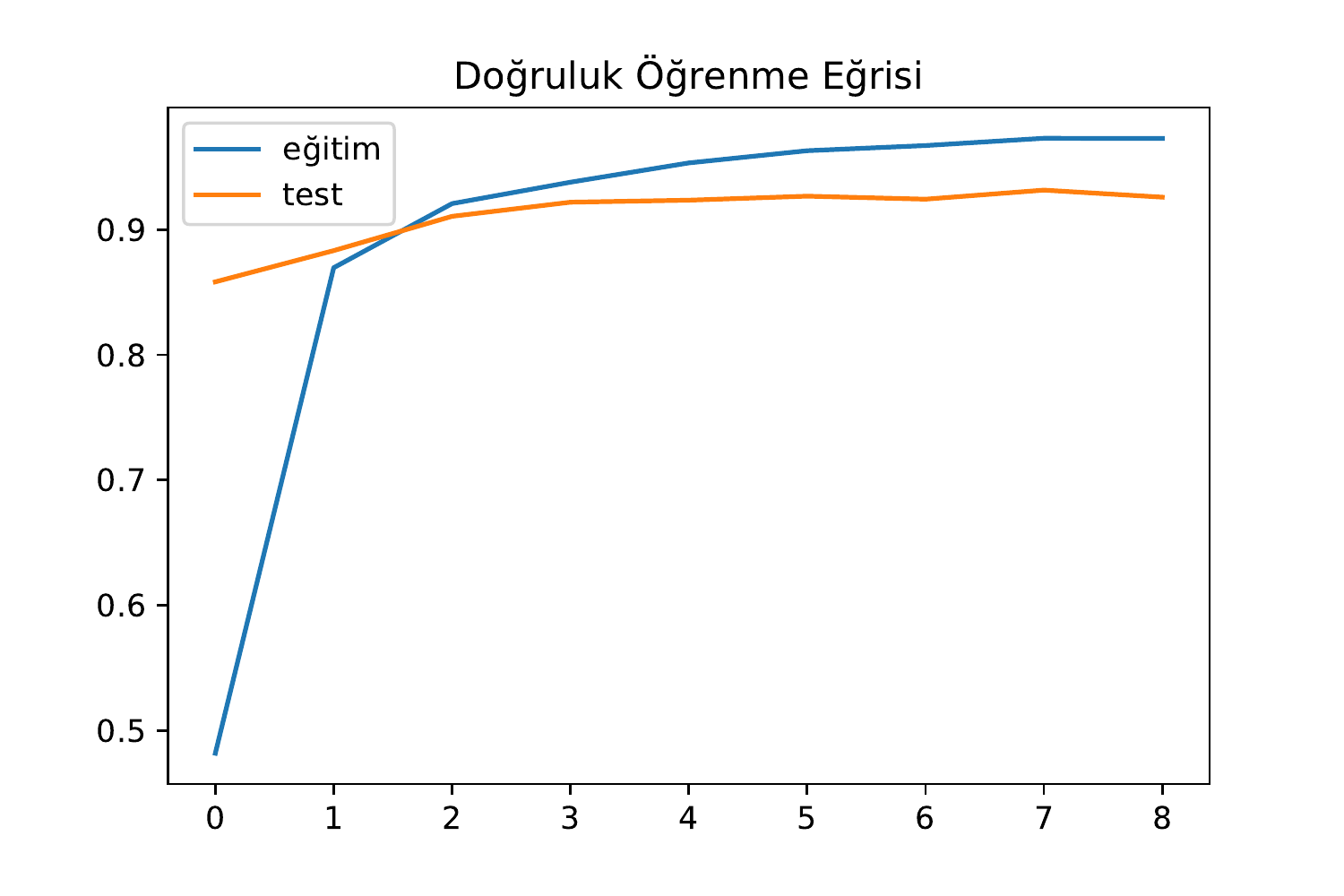}\label{img:lstmAcc}} 
       	     	
	\caption{Bölüm \ref{subsec:lstm}'de anlattığımız LSTM modelinin 10 dönemlik eğitimi sonucunda elde edilen eğriler. \protect\subref{img:lstmLoss} Kayıp değerinin dönem ile değişimini eğitim ve test veri setleri için gösteren eğri \protect\subref{img:lstmAcc} Doğruluk değerinin dönem ile değişimini eğtim ve test veri setleri için gösteren eğri.  } 
	\label{img:lstmTraining}
\end{figure*}

Bölüm \ref{subsec:lstm}'de bahsettiğimiz LSTM yöntemi dahilinde önerilen mimari, ilgili problemin çözümünde Doc2Vec yöntemine göre çok daha iyi oranlarda doğruluk sonuçları vermiştir. Öğrenme eğrilerinin ilk 10 dönem (epoch) için seyri Şekil \ref{img:lstmTraining}'te görülebilir. Çalışmamızda kullanılan diğer yöntemler bu problem özelinde 3 iterasyondan fazla eğitime gerek duymadığı için öğrenme eğrileri sadece bu model için paylaşılmıştır. LSTM modeli için hesaplanan iki F1 skoru için de görece yüksek değerlere ulaşmıştır. Ancak tablodan da görüleceği üzere, daha güncel bir yöntem olan BERT ile kurgulanan ve önceden eğitilmiş bir BERT modelinin ince ayar yapılarak \cite{Schweter:2020} kullanıldığı yöntem en yüksek doğruluk değerine ulaşmıştır. Her ne kadar LSTM yöntemi eğitim veri seti için BERT ile yaklaşık aynı doğruluk değerlerine erişse de, BERT yöntemi test veri setinde daha başarılı olmuştur. Buna bağlı olarak da F1 skorlarında da diğer iki yönteme göre en iyi sonuçları vermiştir. 

Yöntemlerin her bir etiket için başarımlarının ayrı ayrı gösterilmesi adına, F1 skoru hesabında kullanılan hata dizeyleri incelendiğinde, yöntemlerin birbirine göre başarım oranları görsel olarak gözlemlenebilmektedir. Şekil \ref{img:confmxs}'de  her bir yöntem için çizdirdiğimiz hata dizeyleri incelendiğinde, daha yüksek başarımlar daha koyu renklerle gösterildiğinde, BERT yönteminin hata dizeyinin köşegeni doğrultusundaki değerlerin daha yüksek değerlere, dolayısı ile daha koyu renklere sahip olduğu görülmektedir. Tablo \ref{tablo:yontemler}'de gösterdiğimiz ağırlıklı F1 skorlarından da görmüş olduğumuz gibi, BERT yöntemi her bir etiket grubu için en yüksek doğruluk oranına sahip yöntem olmuştur.

\section{Sonuçlar}
Kısa cümlelerin sınıflandırılması, benzer kelimeler içermeyen cümleler için geleneksel yöntemlerle kolaylıkla gerçekleştirilebilmektedir. Çok benzer kelimeler içeren kısa cümlelerin ait olduğu farklı kategorileri doğru şekilde sınıflandırabilmek için derin öğrenme yöntemlerine başvurmanın doğru bir yaklaşım olabileceği tespit edilmiştir. Bu noktada yapılan analizler BERT ile yapılan sınıflandırma işlemlerinin incelenen diğer yöntemlerden daha yüksek doğruluk oranıyla bu işlemi gerçekleştirebildiğini ortaya koymuştur. İlerleyen çalışmalarda aynı modellerin çok daha fazla kategoride ve farklı uzunluklardaki cümleler ve cümle grupları ile performansı karşılaştırılabilir.

Çalışmamızda gerçekleştirdiğimiz kodlama çalışmaları için python dilini ve ilgili kütüphanelerinin yanısıra düzenleme ve derleme için Jupyter Notebook uygulamasını kullandık.  Bu kodlar ilgili GitHub depomuzdan \cite{AdresGezgini:2021} klonlanarak benzer otomatik etiketleme problemleri için kolaylıkla test edilebilir.

\section*{B{\footnotesize İ}lg{\footnotesize İ}lend{\footnotesize İ}rme}

Bu çalışma TÜBİTAK TEYDEB 1501 programı kapsamında desteklenmiş olan 3190585 numaralı "Makine Öğrenmesi ile Anlamlı Diyalog Üretebilecek Genel Amaçlı Chatbot Uygulaması" isimli proje kapsamında gerçekleştirilmiştir.

\bibliographystyle{IEEEbib}
\bibliography{referanslar}

\end{document}